\documentclass[11pt,a4paper]{article}
\usepackage[hyperref]{acl2021}
\usepackage{times}
\usepackage{graphicx}
\usepackage{latexsym}

\usepackage{dsfont}
\usepackage{dirtytalk}
\usepackage{multirow}
\usepackage{subfig}
\usepackage{float}

\usepackage{microtype}

\aclfinalcopy 
\def\aclpaperid{1729} 

\definecolor{ForestGreen}{RGB}{34,139,34}

\title{COVID-Fact: Fact Extraction and Verification of Real-World Claims on COVID-19 Pandemic}

\author{Arkadiy Saakyan\textsuperscript{1}, 
  Tuhin Chakrabarty \textsuperscript{1}, 
  \textbf{and} \textbf{Smaranda Muresan}\textsuperscript{1,2}\\ 
  \textsuperscript{1}Department of Computer Science, Columbia University \\
  \textsuperscript{2}Data Science Institute, Columbia University\\
  {\tt a.saakyan@columbia.edu}, {\tt \{tuhin.chakr,smara\}@cs.columbia.edu}
  }

\date{}

\begin{document}
\maketitle
\begin{abstract}
We introduce a FEVER-like dataset COVID-Fact of $4,086$ claims concerning the COVID-19 pandemic. The dataset contains claims, evidence for the claims, and contradictory claims refuted by the evidence. Unlike previous approaches, we automatically detect true claims and their source articles and then generate counter-claims using automatic methods rather than employing human annotators. Along with our constructed resource, we formally present the task of identifying relevant evidence for the claims and verifying whether the evidence refutes or supports a given claim. In addition to scientific claims,  our data contains simplified general claims from media sources, making it better suited for detecting general misinformation regarding COVID-19.  Our experiments indicate that COVID-Fact will provide a challenging testbed for the development of new systems and our approach will reduce the costs of building domain-specific datasets for detecting misinformation.
\end{abstract}

\section{Introduction}
The proliferation of disinformation and misinformation on the web is increasing at a scale that calls for the  automation of the slow and labor-intensive manual fact-checking process \cite{vosoughi2018spread}. \textit{New York Times} reports that ``Physicians say they regularly treat people more inclined to believe what they read on Facebook than what a medical professional tells them." Disinformation is even more acute around the recent COVID-19 pandemic. As a result, there is a need for automated fact-checking tools to assist professional fact-checkers and the public in evaluating the veracity of claims that are propagated online in news articles or social media. 

Ideally, a fact-checking pipeline will address several tasks: 1) Consider real-world claims, 2) Retrieve relevant documents not bounded to a known document collection (e.g., Wikipedia) and which contain information to validate the claim, 3) Select evidence sentences that can support or refute the claim and 4) Predict the claim veracity based on this evidence. Recent work on end-to-end fact-checking, including models and datasets, has advanced the field by addressing several tasks in the pipeline, but not all \cite{N18-1074,thorne-etal-2019-fever2,hanselowski-etal-2019-richly,augenstein-etal-2019-multifc,diggelmann2021climatefever,wadden-etal-2020-fact}.  One line of work that includes FEVER \cite{N18-1074,thorne-etal-2019-fever2} and SciFact \cite{wadden-etal-2020-fact}
addresses tasks 2, 3 and 4, but assumes a given document collection for task 2 (Wikipedia or CORD-19, respectively) and does not address task 1. Moreover, the refuted claims in these datasets are manually generated by asking humans to produce counter-claims for a given claim supported by a source document. Another line of work that includes Multi-FC \cite{augenstein-etal-2019-multifc} addresses tasks 1, 2 and 4, but not 3. It provides real-world claims  collected from fact-checking websites and evidence documents and other meta-information, but it does not provide evidence sentences.

\begin{figure}[t]
\centering
\includegraphics[scale=0.18]{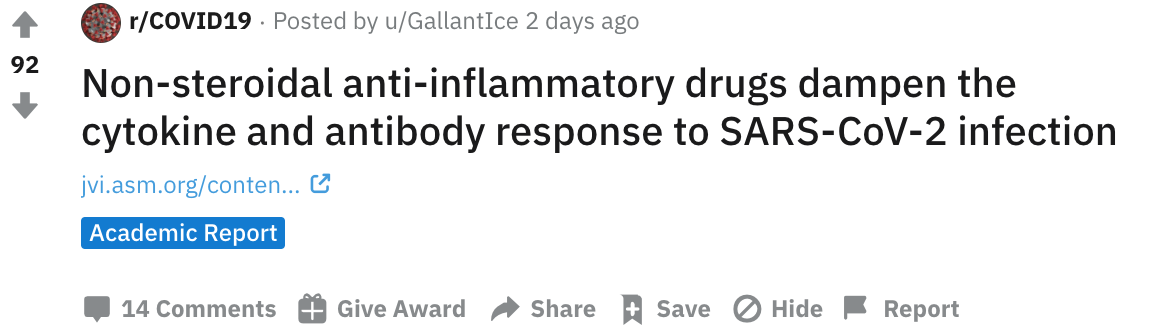}
\caption{\label{figure:image1} A claim from the \textit{r/COVID19} subreddit with an academic report as an evidence source linked to it.}
\end{figure}

\begin{table*}
\renewcommand{\arraystretch}{1.25}
\centering
\small
\begin{tabular}{|l|l|}
\hline
Original Claim    & Closed environments \textbf{\color{ForestGreen}facilitate} secondary transmission of coronavirus disease 2019                                                                                                 \\ \hline
Counter-Claim & Closed environments \textbf{\color{red}prevent} secondary transmission of coronavirus disease 2019                                                                                                    \\ \hline
Gold Document     & \url{https://www.medrxiv.org/content/10.1101/2020.02.28.20029272v2}                                                                                                                     \\ \hline
Gold Evidence     & \begin{tabular}[c]{@{}l@{}}It is plausible that closed environments contribute to secondary transmission of \\ COVID-19 and promote superspreading events.\end{tabular}           \\ \hline\hline
Original Claim    & Oxford vaccine \textbf{\color{ForestGreen}triggers} immune response                                                                                                                                           \\ \hline
Counter-Claim & Oxford vaccine \textbf{\color{red}inhibits} immune response                                                                                                                                           \\ \hline
Gold Document     & \url{https://www.bbc.com/news/uk-53469839}                                                                                                                                              \\ \hline
Gold Evidence     & \begin{tabular}[c]{@{}l@{}}They are injecting coronavirus RNA (its genetic code), which then starts making \\ viral proteins in order to trigger an immune response.\end{tabular} \\ \hline
\end{tabular}
\caption{\label{table:table1}Original and counter-claims from our dataset with gold documents and evidence sentences identified by our system supporting and refuting them, respectively.}
\end{table*}

We propose a novel semi-automatic method to build a fact-checking dataset for COVID-19 (COVID-Fact) with the goal of facilitating all the above tasks. We make the dataset and code available for future research at \url{https://github.com/asaakyan/covidfact}.  Our contributions are as follows: 
\begin{itemize}
\item \textit{Automatic real-world true claim and trustworthy evidence document selection} (Section \ref{section:claim}). We start with the heavily moderated \textit{r/COVID19} subreddit, that requires every claim/title post to be accompanied by a source evidence document from peer-reviewed research, pre-prints from established servers, or information reported by governments and other reputable agencies. 
Figure \ref{figure:image1} shows one such claim with the associated source belonging to the \textit{Academic Report} flair. We propose additional filtering methods to ensure source quality and that claims are well-formed. This step provides us with real-world true claims about COVID-19 and evidence documents not bounded to a known document collection. Moreover, the language of the claims could be both technical and lay (see Figure \ref{figure:image1} and Table \ref{table:table1}), unlike SciFact which is geared only towards scientific claims.  
\item \textit{Automatic generation of counter-claims} (Section \ref{section:counter-claim}). An end-to-end fact-checking system requires both true and false claims for training. Following FEVER and SciFact, to obtain false claims, we aim to generate counter-claims of the original true claim. The advantage is that we obtain evidence documents/sentences for free. However, unlike FEVER and SciFact, we propose a novel approach to automatically generate counter-claims from a given claim using two steps: 1) select salient words from the true claim using attention scores obtained from a BERT \cite{devlin-etal-2019-bert} model fine-tuned on the SciFact dataset, and 2) replace those words with their opposites using Masked Language Model infilling with entailment-based quality control. Table \ref{table:table1} shows examples of generated counter-claims. 
\item \textit{Evidence sentence selection using text similarity and crowdsourcing} (Section \ref{section:evidence}).  For evidence sentence selection, we calculate the semantic similarity between the original true claim and the sentences in source evidence documents using sentence-BERT (SBERT) \cite{reimers-gurevych-2019-sentence}, retrieve top five sentences and use crowdsourcing for final validation. Table \ref{table:table1} shows examples of evidence sentences that support the true claims and refute the corresponding counter-claims. 
\item \textit{COVID-Fact dataset of 4,086 real-world claims annotated with sentence-level evidence and a baseline on this task}. Our results show that models trained on current datasets (FEVER, SciFact) do not perform well on our data (Section \ref{section:results}). Moreover, we show the usefulness of our dataset through zero-shot performance on the scientific claim verification task on SciFact \cite{wadden-etal-2020-fact} data (Section \ref{section:results}).  
\end{itemize}

\section{COVID-Fact Dataset Construction}
The COVID-Fact dataset contains $4,086$ real-world claims with the corresponding evidence documents and evidence sentences to support or refute the claims. 
There are $1,296$ supported claims and $2,790$ automatically generated refuted claims. In this section, we present the three main steps to semi-automatically construct this dataset: \textbf{1)} real-world true claim and trustworthy evidence document selection (Section \ref{section:claim}), \textbf{2)} automatic counter-claim generation (Section \ref{section:counter-claim}) and \textbf{3)} evidence sentence selection (Section \ref{section:evidence}).

\subsection{Real-World Claim and Trustworthy Evidence Document Selection}\label{section:claim}

The subreddit \textit{r/COVID19} is a heavily moderated online discussion forum that seeks to facilitate scientific discussion around COVID-19. Each post on this subreddit has a title and needs to contain a link to a source, governed by several rules: posts linking to non-scientific sources will be removed; comments making a statement as fact or which include figures or predictions also need to be supported by evidence; allowed sources include peer-reviewed research, pre-prints from established servers, and information reported by governments and other reputable agencies. Moreover, the posts are annotated with ``flairs", or short description of the posts' category such as Academic Report, Academic Comment, Preprint, Clinical, Antivirals, Government Agency, Epidemiology, PPE/Mask research, General. Having access to such flairs allows to select claims, for example, related to ``Vaccine research" or ``Epidemiology". This could further help in training models targeting even more specific types of disinformation, like disinformation about antivirals or PPE/masks. In our study, the titles of the post are considered \emph{candidate claims} and the associated sources are considered \textit{evidence documents}. Posts from the \textit{r/COVID19} subreddit are extracted via the Pushshift Reddit API.\footnote{\url{https://github.com/pushshift/api}} Two issues still need to be addressed: 1) ensure that titles are well-formed claims; 2) ensure the highest trustworthiness of the posts and their associated sources.  
\paragraph{Filtering for well-formed claims.} The definition of a claim can vary depending on domain, register or task \cite{daxenberger-etal-2017-essence}. For our work, we consider a claim to be a proposition whose truthfulness can only be determined by additional evidence. In addition, a well-formed claim has to be a full sentence. Thus, to filter out most of the titles that are not well-formed claims, we employ a simple syntax-based approach to remove questions and consider statements that have at least a main verb.   
This filtering steps allows us to remove titles such as "B cell memory: understanding COVID-19" and consider titles such as the ones in Figure \ref{figure:image1} and Table \ref{table:table1}. In addition, we ask three volunteer computer science students with background in argumentation and linguistics to manually verify that the entire resulting set does indeed contain only well-formed claims. While we could have employed 
more sophisticated claim detection methods, there are no large-scale datasets for COVID-19 to train a claim detection model. We therefore did not want to introduce additional noise in our dataset by using a machine learning approach.

\paragraph{Filtering for trustworthiness.} To ensure high trustworthiness of posts (and thus our true claims) and the linked sources, we employ several filtering steps. 
First, the posts in this subreddit undergo moderation, and thus we discard titles/claims that belong to posts flagged as taken down by the moderators using the posts' ``removed" flair. Moreover, users of the Reddit platform may upvote or downvote a post, and the ratio of upvotes can serve as a rough indication of the reliability of the source. Hence, posts (and thus claims/sources) with upvote ratio lower than 0.7 are rejected. 
We then reject claims where the linked source in the post has an Alexa Site Rank\footnote{\url{https://www.alexa.com/topsites}} lower than $50,000$, rejecting the outliers by the site rank (see the box plot in Appendix \ref{appendix:alexa}). Finally, we reject a claim if the linked source in the post does not appear in the top $5$ Google search results when querying the title of the post.


 From an initial set of $22,646$ posts, automatic syntactic filtering for well-formed claims results in a set of $6,154$ claims, further reduced to $1,526$ after filtering for trustworthiness and finally reduced to $1,407$ through manual validation. Thus, the resulted dataset after all the filtering steps consists of $1,407$ true claims and the associated source evidence documents (an additional set of $111$ claims are removed in the evidence sentence selection step in Section \ref{section:evidence}). Besides the linked source document in the post, we retrieve for each claim four additional sources from the top $5$ Google search results. This is motivated by the fact that the same claim can be reported by various sources. For example, the second claim in Table \ref{table:table1} ``\textit{Oxford vaccine triggers immune response}" is reported, besides the \textit{bbc.com} given in the original post, also by other trustworthy sources such as \textit{usnews.com}, \textit{medscape.com}, \textit{cnbc.com}. 
Unlike FEVER and SciFact, which constrain their evidence document collection to Wikipedia or pre-selected scientific articles, we collect evidences from any of the websites linked to the Reddit post or appearing in the top $5$ Google search results. Even though over time the Google search results may change, the collection of evidence documents for COVID-Fact is considered fixed and will be released for reproducibility.

Like SciFact \cite{wadden-etal-2020-fact}, our dataset contains several claims with scientific jargon such as \say{\textit{Altered blood cell traits underlie a major genetic locus of severe COVID-19}}. However, unlike SciFact, our dataset also contains scientific claims expressed in lay terms. 
For example, a claim like \say{\textit{Loss of smell is a symptom of COVID-19}} is much simpler and can be understood by a wider audience compared to  \say{\textit{Emerging evidence supports recently acquired anosmia and hyposmia as symptoms of COVID-19}}. This is important, as a lot of (dis)information is expressed in lay language intended for the general public not versed in scientific language. Another issue adding to the complexity of the task around COVID-19 (dis)information are non-scientific claims that focus on public health policies or statements from public health authorities. For example, a claim like ``\textit{CDC says new COVID strain in UK could already be circulating undetected in U.S}" would not occur in scientific literature, but occurs in media outlets linked as sources in the \textit{r/COVID19} subreddit.

\subsection{Automatic Counter-Claim Generation}\label{section:counter-claim}

An end-to-end fact-checking system requires both true and false claims. Following FEVER and SciFact, to obtain false claims we aim to generate counter-claims of the original true claims (from Section \ref{section:claim}). However, in FEVER \cite{N18-1074} and SciFact \cite{wadden-etal-2020-fact} the generation of counter-claims was done manually by human annotators, which is an expensive approach that might not scale well. We propose an approach to generate counter-claims automatically (see Table \ref{table:table1} for examples). Our counter-claim generation consists of two stages: 1) select salient words from the true claims, and 2) replace those words with their opposite using Mask Language Model infilling with entailment-based quality control. We discuss these steps below. \medskip

\subsubsection{Salient Words Selection} \label{sec:keyword} Salient words (keywords) are essential to the overall semantics of a sentence. For example, in the claim "\textit{Oxford vaccine triggers immune response}", a salient word would be \textbf{``triggers"}.
By changing the word "triggers" to "inhibits"  we change the meaning of above claim to its opposite (counter-claim). 
Recently \citet{zhang-etal-2020-pointer} used YAKE \cite{campos2018yake,CAMPOS2020257}, an unsupervised automatic keyword extraction method for selecting salient words to guide their text generation process. For selecting salient words from a claim, we experiment with YAKE as one of our methods. In addition, we explore an attention-based method described below.

\paragraph{Attention-Based Salience.} Recently, \citet{sudhakar-etal-2019-transforming} use self-attention scores from BERT \cite{devlin-etal-2019-bert} to delete keywords from an input sequence for the task of Style Transfer. They use a novel method to extract a specific attention head and layer combination that encodes style information and that can be directly used as importance scores. Inspired by them, we use the same approach for our task. We fine-tuned BERT for a sentence classification task (veracity prediction) on the SciFact 
\cite{wadden-etal-2020-fact} dataset, and extract the attention scores from the resulting model. Given the SciFact dataset $D ={(x_{1}, y_{1}), ...,(x_{m}, y_{m})}$ where $x_{i}$ is a claim and $y_{i} \epsilon $ \{SUPPORTED, REFUTED\} is a veracity label  we observe that the self-attention based classifier defines a probability distribution over labels:
$p(y|x) = g(v, \alpha)$  where $v$ is a tensor such that $v[i]$ is an encoding of $x[i]$, and $\alpha$ is a tensor of attention weights such that $\alpha[i]$ is the weight attributed to $v[i]$ by the classifier in deciding probabilities for each $y_{j}$ .
The $\alpha$ scores can be treated as importance scores and be
used to identify salient words.

\paragraph{Quality of Salient Words Selection.}
We evaluate how well our salient word selection methods correlate with human judgement.
We randomly select $150$ original claims for an Amazon Mechanical Turk task. The annotators were asked to select a word that could potentially invert the meaning of the sentence if it were to be replaced. For every claim, three separate annotators were recruited which means that we would have at most three different chosen salient keywords. For each claim, we compute the set intersections between the three keywords selected by our automatic methods (YAKE and Attention-based) vs. the keywords selected by the annotators on AMTurk. We found that keywords selected using self-attention scores have a significantly  higher recall (Two-Proportion Z-test with p-value $< .00001$) than YAKE ($68\%$ vs. $54\%$). The average number of words per claim in COVID-Fact is $14$, so the task of selecting one salient keyword is challenging even for humans. Given this, our Recall@3 scores demonstrate the reliability of automatic attention-based salient word selection.

\subsubsection{Masked Language Model Infilling with Entailment-based Quality Control} 
After selecting salient words from the true claims for replacement, we need to provide only paraphrases that are opposite in meaning and consider the context in which these words occur. Language models have been used previously for infilling tasks \cite{donahue-etal-2020-enabling} and have also been used for automatic claim mutation in fact checking \cite{jiang-etal-2020-hover}. Inspired by these approaches, we use the Masked Language Model (MLM) RoBERTa \cite{DBLP:journals/corr/abs-1907-11692} fine-tuned on CORD-19 \cite{wang-etal-2020-cord} for infilling. The fine-tuned RoBERTa is available on Huggingface \footnote{\url{https://huggingface.co/amoux/roberta-cord19-1M7k}}. We generate a large number (10-30) of candidate counter-claims with replaced keywords per each original claim. 

After generating multiple candidate counter-claims based on MLM infilling, we select the ones that have the highest contradiction score with the original claim. To compute the contradiction score we use the RoBERTa \cite{DBLP:journals/corr/abs-1907-11692} model trained on Multi-NLI \cite{N18-1101} due to its size and diversity. The scores are in the range from $0$ to $1$. We first set the minimum score threshold and then select top three claims above the threshold. 

To select the right threshold for contradiction score-based filtering we perform the following experiment. We presented $150$ randomly selected claims to Amazon Mechanical Turk workers. Annotators were presented with the original claim and five generated candidate counter-claims from MLM infilling. They were then asked if those claims are implied by the original claim (hence, for example, noun shifts would be judged as ``not implied"). We labeled claims as \say{contradictory} if the majority of the annotators agreed on the label. We observed a point-biserial correlation of $0.47$ between dichotomous human judgement and continuous contradiction scores, indicating moderate agreement. We convert the contradiction scores to binary outcomes, assigning 1 if the score is above the threshold and 0 otherwise. We compute precision, recall, F1 score and accuracy for different thresholds. As threshold value increases, we see a steady increase in precision, indicating that taking a higher threshold value we are almost guaranteed to select a contradictory sentence (for example, for a threshold of $0.995$, precision is $93\%$). Obviously, this comes at a cost of decreased recall. We selected a threshold of $0.9$ (precision $76\%$), since we want to prioritize precision, but do not want to reduce our dataset too much due to the low recall. At this threshold, our $1,407$ claims generate additional $4,042$ false claims. 
An alternative approach of replacing salient words with antonyms from standard lexicons like WordNet \cite{miller1995wordnet} was considered. However, a suitable antonym was absent in several cases, most notably nouns. The RoBERTa model is able to provide domain-aware substitutions. For example, replacing the word ``humans" by the word ``mice" reverses the meaning of the claim the domain of clinical trial reports, yet the words human and mouse can hardly be considered antonyms. Lexical replacement without consideration of context can also cause grammatical issues. 

Our method of counter-claim generation only changes a single word or a multi-word expression, since pre-trained MLMs like BERT \cite{devlin-etal-2019-bert} and RoBERTa \cite{DBLP:journals/corr/abs-1907-11692} do not allow for multiple word masking. However, this method can be extended to masking multiple words using recent pre-trained language models like BART \cite{lewis-etal-2020-bart}.

\subsubsection{Analysis of Counter-claim Generation}
\begin{figure}[!ht]
\centering
\includegraphics[scale=0.2]{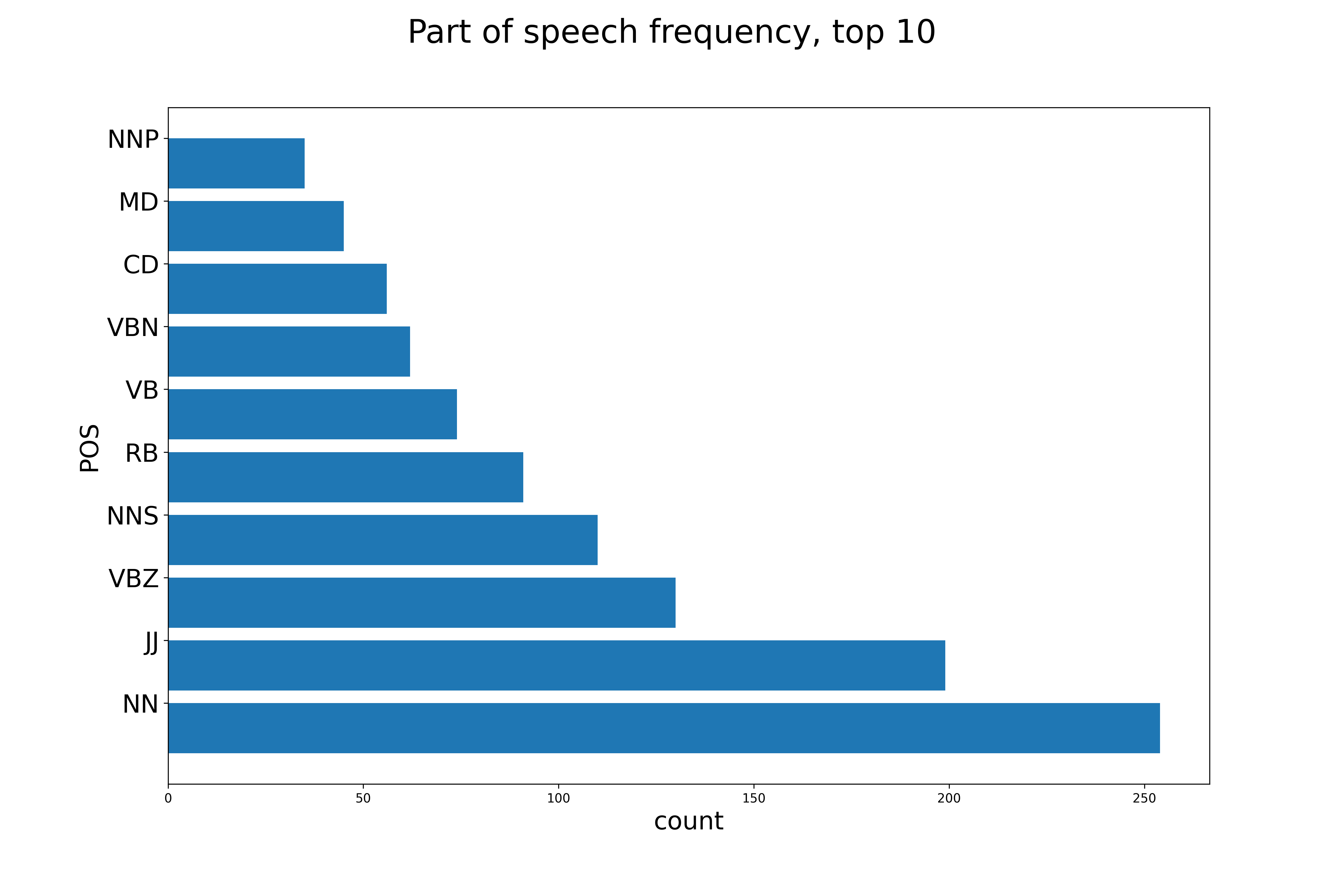}
\caption{Most frequent POS tags of salient words.}
\label{figure:POSog} 
\end{figure}

Upon deeper inspection we observe that the attention scores described in Section \ref{sec:keyword} were distributed 
across different parts of speech like \textit{verbs} or \textit{adjective modifiers} or  \textit{nouns}. 
We show the distribution of the most frequent parts of speech of salient words and replacement words in our dataset in Figure \ref{figure:POSog}. 
This means our counter-claims were generated with more creativity than just the addition of obvious triggers like ``not”. The majority of claim negations involved a reversal of effect direction; for instance ``\textit{Suspicions grow that nanoparticles in Pfizer's COVID-19 vaccine trigger \underline{rare} allergic reactions.}” was negated as  “\textit{Suspicions grow that nanoparticles in Pfizer's COVID-19 vaccine trigger \textbf{{\color{red}systemic}} allergic reactions.}” where a simple adjective modifier changes the truthfulness. Similarly for a claim ``\textit{Electrostatic spraying will \underline{prevent} the spread of COVID-19}" a negated claim is ``\textit{Electrostatic spraying will \textbf{{\color{red}facilitate}} the spread of COVID-19}" which flips the main verb in the claim. In Table $\ref{table:comparison}$, one can see several examples of how the generated counter-claims reverse the meaning of the original sentence.

\begin{table}
\renewcommand{\arraystretch}{1.25}
\small
\centering
\begin{tabular}{|l|l|}
\hline
\multicolumn{1}{|c|}{\textbf{Original}} & \multicolumn{1}{c|}{\textbf{Generated}} \\ \hline
\textit{..\textbf{\color{ForestGreen}people} in UK receive ..} & \textit{..\textbf{\color{red}mice} in UK receive \dots}   \\ \hline
\textit{.. \textbf{\color{ForestGreen}human} ACE2.}       & \textit{.. \textbf{\color{red}bat} ACE2.}          \\ \hline
\textit{\textbf{\color{ForestGreen}FDA} takes key action ..}            & \textit{\textbf{\color{red}WHO} takes key action ..}            \\ \hline\hline
\textit{..\textbf{\color{ForestGreen}improves} the effect ..}             & \textit{..\textbf{\color{red}inhibits} the effect ..}      \\ \hline
\textit{..\textbf{\color{ForestGreen}blocks} SARS-CoV-2..}               & \textit{..\textbf{\color{red}enchanced} SARS-CoV-2..}             \\ \hline\hline
\textit{..are \textbf{\color{ForestGreen}not} fit for purpose ..}   & \textit{..are \textbf{\color{red}good} fit for ..}   \\ \hline
\textit{.. the \textbf{\color{ForestGreen}final} stage ..} & \textit{.. the \textbf{\color{red}first} stage ..}  \\ \hline
\textit{.. shows \textbf{\color{ForestGreen}positive} results.}    & \textit{.. shows \textbf{\color{red}no} results.}           \\ \hline
\end{tabular}
\caption{\label{table:comparison} A detailed look into what parts of speech are replaced, and in what direction the claims are reversed. We omitted full claims due to space constraint. The first 3 claims show nouns, the next 2 show verbs and the final 3 show adjective modifications.}
\end{table}

\begin{table}
\small
\centering
\begin{tabular}{|l|c|c|}
\hline
\textbf{Split} & \multicolumn{1}{l|}{\textbf{Supported}} & \multicolumn{1}{l|}{\textbf{Refuted}} \\ \hline
\textbf{Train} & 1036 & 2227 \\ \hline
\textbf{Dev}   & 130  & 289  \\ \hline
\textbf{Test}  & 130  & 274  \\ \hline
\end{tabular}
\caption{Breakdown of claims by label for train, dev, test sets.}
\label{table:datastat}
\end{table}

\subsection{Evidence Sentence Selection}  \label{section:evidence}

To select evidence sentences we follow the approach proposed by \citet{hidey-etal-2020-deseption}. Given the true claims and the $5$ evidence documents for each claim (Section \ref{section:claim}) we use cosine similarity on SBERT sentence embeddings \cite{reimers-gurevych-2019-sentence} to extract the top $5$ sentences most similar to the true claim. Note that we only need to do this step for true claims, as automatically the evidence sentences that support the true claim will be the evidence sentences that refute the corresponding counter-claims. Sentences containing the claim itself were discarded. The collected five sentences will serve as candidate evidence sentences for future human validation described below. 
\paragraph{Crowdsourcing for Final Evidence Sentence Selection.} Amazon Mechanical Turk workers were given a claim and the $5$ automatically selected candidate evidence sentences. They were asked to select which of the evidence sentences support the claim (they could select several) or they could select that the evidence is absent.
To discourage low quality responses, we used a trick sentence that would allow us to disqualify dishonest entries. For the trick we used a phrase ``\textit{It is not true that}" concatenated with the original sentence, and rejected entries that marked that option as evidence for the claim. In 111 cases, annotators could not agree on the evidence or agreed that the evidence was absent, where agreement is defined as the majority vote. We disregard these true claims from our COVID-Fact dataset as they would not have associated evidence sentences. 

We asses the quality of the majority vote annotations by comparing the gold evidence label annotations with an independent re-annotation by three Amazon Mechanical Turk workers. We select a sample of $100$ claims' evidences ($7\%$ of the $1,296$ original claims). %
We observe a Cohen's kappa \cite{cohen1968weighted} of $0.5$ between majority votes of the two independent groups of Amazon Turk workers, indicating moderate agreement \cite{10.1162/coli.07-034-R2}. We find this encouraging given the complexity of the task, especially considering that the workers did not have domain-specific knowledge.

\section{Experimental setup}


Table \ref{table:datastat} shows the dataset statistics for train/dev/test split of COVID-Fact.

\subsection{COVID-Fact Task Formulation} 
\label{sec:task}
COVID-Fact Task follows the FEVER shared task definition.  The set of all claims is denoted by $C$. The set of gold evidence sentences for a claim $c \in C$ is denoted by $E(c)$. The gold label for a given claim and evidence pair is defined as $v(c, E(c)) \in \{$SUPPORTED, REFUTED$\}$. The task consists of the following subtasks outlined below.

\paragraph{Evidence Retrieval.} Given a claim $c$, a system must retrieve a set of up to five evidence sentences $\hat{E}(c)$. We evaluate the evidence retrieval system quality using precision, recall, and F1 scores. Evidence recall is computed as the number of evidence sets that contain a gold evidence over the total number of evidence sets. 

\paragraph{Veracity Prediction.} Given a claim $c$ and a set of evidence sentences $E(c)$, a system must determine a label $\hat{v}(c, E(c)) \in \{$SUPPORTED, REFUTED$\}$. We evaluate veracity prediction using F1 score and accuracy.  

\paragraph{Evidence Retrieval + Veracity Prediction (COVID-FEVER Score)} Given a claim $c$, a system must retrieve a set of evidence sentences $\hat{E}(c)$, and determine a label $\hat{v}(c, \hat{E}(c)) \in \{$SUPPORTED, REFUTED$\}$. A claim has a COVID-FEVER of 1 if it correctly predicts the veracity of the claim-evidence pair and if at least one of the predicted evidence from the predicted evidence matches the gold evidence selected by annotators (thus a stricter score than veracity prediction accuracy). This metric is similar to the FEVER score \cite{N18-1074}. 

\subsection{Baseline Pipeline for COVID-Fact}
Our end-to-end pipeline consists of the following steps: 1) Evidence retrieval using Google Search + SBERT 
2) Veracity prediction using RoBERTa fine-tuned on fact-checking and entailment inference datasets.

\paragraph{Baseline for Evidence Retrieval.} We use the same approach as was used for the construction of the dataset to provide a strong baseline for evidence retrieval on COVID-Fact. Google search was used to identify five potential source documents by querying the claim.
This step is followed by selecting most similar sentences through computing cosine similarity between sentence embeddings of the claim and candidate sentences using SBERT  \cite{reimers-gurevych-2019-sentence}. 

\paragraph{Baseline for Veracity Prediction.} Our baseline for veracity prediction is a  RoBERTa model. We concatenate all evidence sentences in the evidence set and use it as input for a binary classification task similar to the GLUE RTE task \cite{wang-etal-2018-glue}. We evaluate the models with gold evidence, as well as Top-$5$ and Top-$1$ evidences ranked by SBERT cosine similarity with the original claim.

\subsection{Experiments}
Besides evaluating our baseline pipeline on the COVID-Fact dataset, we perform several additional experiments outlined below. All hyperparameters can be found in Appendix \ref{appendix:model}. 

\paragraph{Adequacy of Existing Datasets for COVID-Fact.}  
For the task of veracity prediction, we evaluate the performance of RoBERTa-large fine-tuned on FEVER, SciFact, MNLI and our COVID-Fact dataset. Moreover, we also experiment with fine-tuning RoBERTa-large on SciFact + COVID-Fact and on FEVER + COVID-Fact. 

\paragraph{Usefulness of COVID-Fact for Zero-Shot Scientific Fact-checking.} 
Even though not explicitly designed for COVID-19 related claims, \citet{wadden-etal-2020-fact} showed how models trained on the SciFact dataset could verify claims about COVID-19 against the research literature. COVID-Fact on the contrary was not explicitly designed for scientific fact-checking, although our resource contains a substantial number of scientific claims. This provides us the opportunity to test the generalizability and robustness of our dataset. To do so, we train models on COVID-Fact claims and gold evidence and evaluate the veracity performance on the SciFact dev set in a zero-shot setting. We remove the NOT ENOUGH INFO claims from the SciFact dataset.

\begin{table*}
\small
\centering
\begin{tabular}{|c|c|c|c|c|c|c|c|c|c|c|c|}
\hline
\multicolumn{9}{|c|}{Veracity Prediction}                                                                                                                           &  & \multicolumn{2}{l|}{COVID-FEVER} \\ \hline
                                                      & \multicolumn{3}{c|}{Gold}             & \multicolumn{3}{c|}{Top 5}             & \multicolumn{2}{c|}{Top 1} &  & \multicolumn{2}{l|}{Top 5}       \\ \cline{2-12} 
                                                      & \multicolumn{2}{c|}{Acc}  & F1        & Acc        & \multicolumn{2}{c|}{F1}   & Acc          & F1          &  & \multicolumn{2}{l|}{Score}       \\ \hline
MNLI \cite{N18-1101}                 & \multicolumn{2}{c|}{61.3} & 64.2      & 53.1       & \multicolumn{2}{c|}{51.5} & 65.4         & 60.6        &  & \multicolumn{2}{l|}{35.1}            \\ \hline
SciFact \cite{wadden-etal-2020-fact} & \multicolumn{2}{c|}{56.9} & 57.0      & 53.7       & \multicolumn{2}{c|}{54.0} & 54.3         & 54.0        &  & \multicolumn{2}{l|}{36.9}            \\ \hline
FEVER \cite{N18-1074}                & \multicolumn{2}{c|}{48.3} & 47.0      & 46.2       & \multicolumn{2}{c|}{45.0} & 48.6         & 48.0        &  & \multicolumn{2}{l|}{35.4}            \\ \hline\hline
COVID-Fact                                            & \multicolumn{2}{c|}{\textbf{83.5}} & \textbf{82.0}     & \textbf{84.7}       & \multicolumn{2}{c|}{\textbf{83.0}} & \textbf{\textbf{83.2}}         & \textbf{81.0}        &  & \multicolumn{2}{l|}{\textbf{43.3}}        \\ \hline
SciFact + COVID-Fact                                  & \multicolumn{2}{c|}{82.2} & 81.0      & 83.0       & \multicolumn{2}{c|}{82.0} & 80.2         & 79.0        &  & \multicolumn{2}{l|}{43.0}            \\ \hline
FEVER + COVID-Fact                                    & \multicolumn{2}{c|}{74.8} & 70.0      & 78.2       & \multicolumn{2}{c|}{73.0} & 73.3         & 68.0        &  & \multicolumn{2}{l|}{35.4}            \\ \hline\hline
COVID-Fact (Claim only)                               & \multicolumn{2}{c|}{67.5} & 40.0      & -          & \multicolumn{2}{c|}{-}    & -            & -           &  & \multicolumn{2}{l|}{-}           \\ \hline
\end{tabular}
\caption{\label{table:veracity} Performance of various training configurations of RoBERTa-large in the Veracity Prediction as well as Evidence Retrieval + Veracity Prediction (See Section \ref{sec:task}). The top 3 rows under Veracity Prediction show a zero shot setting where models are trained on existing fact-checking datasets. We test the model performance on claims with gold evidence selected by humans VS claims with top 5 retrieved evidences and top 1 retrieved evidence on COVID-Fact test set. $p<.001$ using approximate randomization test.}
\end{table*}

\begin{table}[!ht]
\small
\centering
\begin{tabular}{|l|c|l|c|c|}
\hline
\multirow{2}{*}{\textbf{}} & \multicolumn{4}{l|}{\textbf{Evidence Retrieval}}                           \\ \cline{2-5} 
                           & \multicolumn{2}{c|}{P}              & R                 & F1               \\ \hline
Top 5                      & \multicolumn{2}{c|}{22.27}          & \textbf{52.37}    & 31.25            \\ \hline
Top 3                      & \multicolumn{2}{c|}{24.77}          & 45.14             & \textbf{31.99}   \\ \hline
Top 1                      & \multicolumn{2}{c|}{\textbf{29.68}} & 29.93             & 29.80            \\ \hline
\end{tabular}
\caption{\label{table:evidence}Performance of our system's Evidence Retrieval part  (see Section  \ref{sec:task}). We compare the precision (P), Recall (R), and F1-score of top 5, top 3, top 1 retrieved sentences, ranked by SBERT cosine similarity score.}
\end{table}

\section{Results and Analysis} \label{section:results}

Table \ref{table:evidence} summarizes the results for the evidence retrieval evaluation. Our pipeline provides a strong baseline with F1 score of $\approx 32$. For comparison, the baseline system in FEVER \cite{thorne-etal-2019-fever2} achieves the F1 score of $18.26$. Note Top $5$ evidence retrieval performs worse than gold since we evaluate how the system performs with automatically negated claims as well, for which we re-run the Google+SBERT method. 

Table \ref{table:veracity} summarizes the results for the veracity prediction task using gold and retrieved evidence. We observe that, given the gold evidences, fine-tuning on COVID-Fact led to performance improvement of $25$ F1-score and $35$ F1-score compared to training solely on SciFact and FEVER respectively. This indicates that the COVID-Fact dataset is challenging and cannot be solved using popular fact-checking datasets like FEVER and SciFact. This could be explained by the fact that claims about COVID-19 are comprised of a mix of scientific and general-domain claims. The poor macro-F1 score for claim only baseline shows that the model does not learn spurious correlation between a claim and the veracity label. With Top $5$ and Top $1$ retrieved evidences, we observed that COVID-Fact is still difficult to outperform. The zero-shot performance is negligibly affected by the retrieved evidence. Our baseline pipeline achieves the COVID-FEVER score of 43.3 using Top $5$ evidence sentences. Adding the FEVER and SciFact datasets deteriorates the results. 

Table \ref{table:zeroshot} shows a strong zero shot performance of COVID-Fact for scientific claim verification (training on COVID-Fact train set, testing on the SciFact dev set). SciFact only contains scientific claims, therefore the model trained only on SciFact does not generalize well to COVID-Fact, which also contains non-scientific claims. COVID-Fact, on the other hand, contains  enough scientific claims so that the model generalizes well to SciFact. This result shows semi-automated COVID-Fact is not inferior to mostly manual SciFact.

\begin{table}
\small
\centering
\begin{tabular}{|l|l|l|}
\hline
   Train Setting        & Acc  & F1   \\ \hline
COVID-Fact & 80.8 & 80.0 \\ \hline
Sci-Fact   & 83.7 & 83.0 \\ \hline
\end{tabular}
\caption{\label{table:zeroshot} Two-way Veracity prediction results on Sci-fact dev set by models trained on COVID-Fact data as well as Sci-Fact data.}
\end{table}

\paragraph{Error analysis.} We observe that errors in veracity prediction can be attributed to three factors: Cause and Effect, Commonsense or Scientific Background. For instance, in the first (C1, EV1) pair in Table \ref{table:table6},  \textit{not detectable} is the Cause while \textit{testing negative} is the Effect. To verify this claim, the veracity model needs to have knowledge of counterfactuals. Furthermore, it should be understood that \textit{All 10 patients} mention in EV1 should refer to  \textit{women} in C1, due to mention of ``vaginal fluids" but this requires commonsense knowledge outside the text. Finally, it might be hard for veracity models to correctly classify claim evidence pairs which include knowledge of domain-specific or scientific lexical relationships. For instance in (C2, EV2) we see that both bolded phrases in red and blue refer to the same phenomena, but \textit{immune dysregulation} is ``a breakdown of immune system processes" and restraining it can be seen as the same concept as \textit{correcting immune abormalities}, but the model is not able to capture such complex domain specific knowledge.

\begin{table}
\renewcommand{\arraystretch}{1.25}
\small
\centering
\begin{tabular}{|l|l|}
\hline
C1 & \begin{tabular}[c]{@{}l@{}}SARS-CoV-2 is \textbf{\color{red}not detectable} in the vaginal\\ fluid of \textbf{\color{red}women} with severe COVID-19 infection\end{tabular}                                                                                     \\ \hline
EV1   & \begin{tabular}[c]{@{}l@{}}\textbf{\color{blue}All 10 patients} were tested for SARS-CoV-2 in \\ vaginal fluid,and all samples tested \textbf{\color{blue}negative} for\\  the virus.\end{tabular}                                                                 \\ \hline\hline
C2 & \begin{tabular}[c]{@{}l@{}}Baricitinib \textbf{\color{red}restrains the immune }\textbf{\color{red}dysregulation}\\ in COVID-19 patients\end{tabular}                                                                                                            \\ \hline
EV2  & \begin{tabular}[c]{@{}l@{}}Here, we provide evidences on the efficacy of \\Baricitini, a JAK1/JAK2 inhibitor, in  \textbf{\color{blue}correcting}\\ \textbf{\color{blue} the immune abnormalities} observed in patients \\hospitalized with COVID-19.\end{tabular} \\ \hline
\end{tabular}
\caption{\label{table:table6} Claims (C1 \& C2) which are classified incorrectly as REFUTES in the light of SUPPORTing evidence by our best veracity models. Words crucial for correct verification are highlighted.}

\end{table}

\section{Related Work}

\paragraph{Fact-Checking.}
Approaches for predicting the veracity of naturally-occurring claims have focused on statements fact-checked by journalists or organizations such as  PolitiFact.org \cite{W14-2508,alhindi-etal-2018-evidence}, news articles \cite{raofakenews}, or answers in community forums \cite{DBLP:journals/corr/abs-1803-03178, mihaylova2019semeval}. Mixed-domain large scale datasets such as UKP Snopes \cite{hanselowski-etal-2019-richly}, MultiFC \cite{augenstein-etal-2019-multifc}, and FEVER \cite{N18-1074,thorne-etal-2019-fever2} rely on Wikipedia and fact-checking websites to obtain evidences for their claims. Even though these datasets contain many claims, due do domain mismatch they may be difficult to apply for COVID-19 related misinformation detection. SciFact \cite{wadden-etal-2020-fact} introduced the task of scientific fact-checking, generating a dataset of 1.4K scientific claims and corresponding evidences from paper abstracts annotated by experts. However, the dataset does not contain simplified scientific claims encountered in news and social media sources, making it difficult to optimize for a misinformation detection objective. Another approach to misinformation detection similar to ours is CLIMATE-FEVER \cite{diggelmann2021climatefever}. They adapted FEVER methodology to create a dataset specific to climate change fact-checking. However, due to difficult and expensive methods employed for generation of FEVER, it can be difficult to extrapolate this method to assemble a COVID-19 specific dataset.

\paragraph{COVID-19 related NLP tasks.}
Numerous NLP approaches were employed to aid the battle with the COVID-19 pandemic. Notably \citet{wang-etal-2020-cord} released CORD-19, a dataset containing 140K papers about COVID-19 and related topics while \citet{zhang-etal-2020-rapidly} created a neural search engine COVIDEX for information retrieval. To combat misinformation \citet{lee2020misinformation} proposed a hypothesis that misinformation has high perplexity. \citet{hossain-etal-2020-COVIDlies} released COVIDLIES: a dataset of 6761 expert-annotated tweets matched with their stance on known COVID-19 misconceptions. The dataset provides a comprehensive evaluation of misconception retrieval but does not analyze evidence retrieval and prediction of veracity of claims based on presented evidence. \citet{poliak-etal-2020-collecting} collected 24,000 Question with expert answers from 40 trusted websites to help NLP research with COVID related information. COVID-Fact, on the other hand, deals with real world claims and presents an end-to-end fact checking system to fight misinformation.

\section{Conclusion}
We release a dataset of 4,086 claims concerning the COVID-19 pandemic, together with supporting and refuting evidence. The dataset contains real-world true claims obtained from the \textit{r/COVID19} subreddit as well as automatically generated  counter-claims. Our experiments reveal that our dataset outperforms zero-shot baselines trained on popular fact-checking benchmarks like SciFact and FEVER. This goes on to prove how domain-specific vocabulary may negatively impact the performance of popular NLP benchmarks. Finally, we demonstrate a simple, scalable, and cost-efficient way to automatically generate  counter-claims, thereby aiding in creation of domain-specific fact-checking datasets. We provide a detailed evaluation of the COVID-Fact task and hope that our dataset serves as a challenging testbed for end-to-end fact-checking around COVID-19. 

\section{Ethics}
The data was collected from Reddit keeping user privacy in mind. Reddit is a platform where users post publicly and anonymously. For our dataset, only titles and links to external publicly available sources like news outlets or research journals were collected, as well as post metadata such as flairs, upvote ratio, and date of the post. User-identifying information, including, but not limited to, user's name, health, financial status, racial or ethnic origin, religious or philosophical affiliation or beliefs, sexual orientation, trade union membership, alleged or actual commission of crime, was not retrieved and is not part of our dataset. 
For all the crowdsourcing annotation work, we fairly compensate crowd workers in accordance with local minimum wage guidelines.

One significant concern might arise regarding the use of language models for  counter-claim generation. Our model is a controlled generation system (word-level replacement) and is not suited for generation of entirely new and original claims. Neither it is the case that it can be used for generation of entire articles of false information, or generating false evidence for the counter-claims. The model for replacing keywords from original claims is trained on CORD-19 \cite{wang-etal-2020-cord}, a scientific corpus of high quality and trustworthy information about COVID-19. We generate counter-claims to create a resource that will help NLP models learn how to identify false information and provide evidence for the predicted label leading to more explainable models. Consequently, our approach is suited for improving entailment and veracity prediction performance of fact-checking systems, rather than improving generative qualities of false-claim generation systems. The fact that we use our model to generate false claims also helps to address the concerns of biased language generation. In the unlikely event our model produces biased claims, they could serve as good examples of false claims containing bias, which would be an interesting topic for further research (bias in disinformation). We therefore believe the net positive impact of our work far outweighs the potential risks.



\bibliographystyle{acl_natbib}
\bibliography{anthology,acl2021}
\newpage
\vfill\eject
\clearpage
\appendix
\section{Model implementation details}
\label{appendix:model}
We used fairseq library \cite{ott-etal-2019-fairseq} for RoBERTa model training.
\subsection{Salient word selection hyperparameters}
We use the uncased BERT model since many titles contain words that are all capitalized. We train the model on the SciFact classification task using $15$ epochs and batch size of $16$. The training loss is $7.15e-03$. The rest of the parameters are set to default as in \cite{sudhakar-etal-2019-transforming}.

\subsection{Veracity prediction hyperparameters}
\begin{itemize}
    \item{\textbf{No of Parameters:}} We use the RoBERTA-large checkpoint (355M parameters) and use the FAIRSEQ implementation \cite{ott2019}
     \footnote{\url{https://github.com/pytorch/fairseq/tree/master/examples/roberta}}.
    \item{\textbf{No of Epochs:}} We fine-tune pre-trained RoBERTa for 10 epochs for each model and save the best model based on validation accuracy on COVIDFact. 
    \item{\textbf{Training Time:}} Our training time is 30 minutes for each model except for ones with FEVER which takes around 10 hours.
    \item{\textbf{Hardware Configuration:}} We use 2 RTX 2080 GPUs.
    \item{\textbf{Training Hyper parameters:}} We use the same parameters as the FAIRSEQ github repository where RoBERTa was fine-tuned for the RTE task in GLUE with the exception of the size of each mini-batch, in terms of the number of tokens, for which we used 1024. \footnote{\url{https://github.com/pytorch/fairseq/blob/master/examples/roberta/README.glue.md}}
\end{itemize}
\newpage
\section{Dataset statistics}
Figures below visualize most frequent flairs in the dataset, as well as word clouds with keywords and replaced words.


\begin{figure}[htbp]
\centering
\includegraphics[scale=0.23]{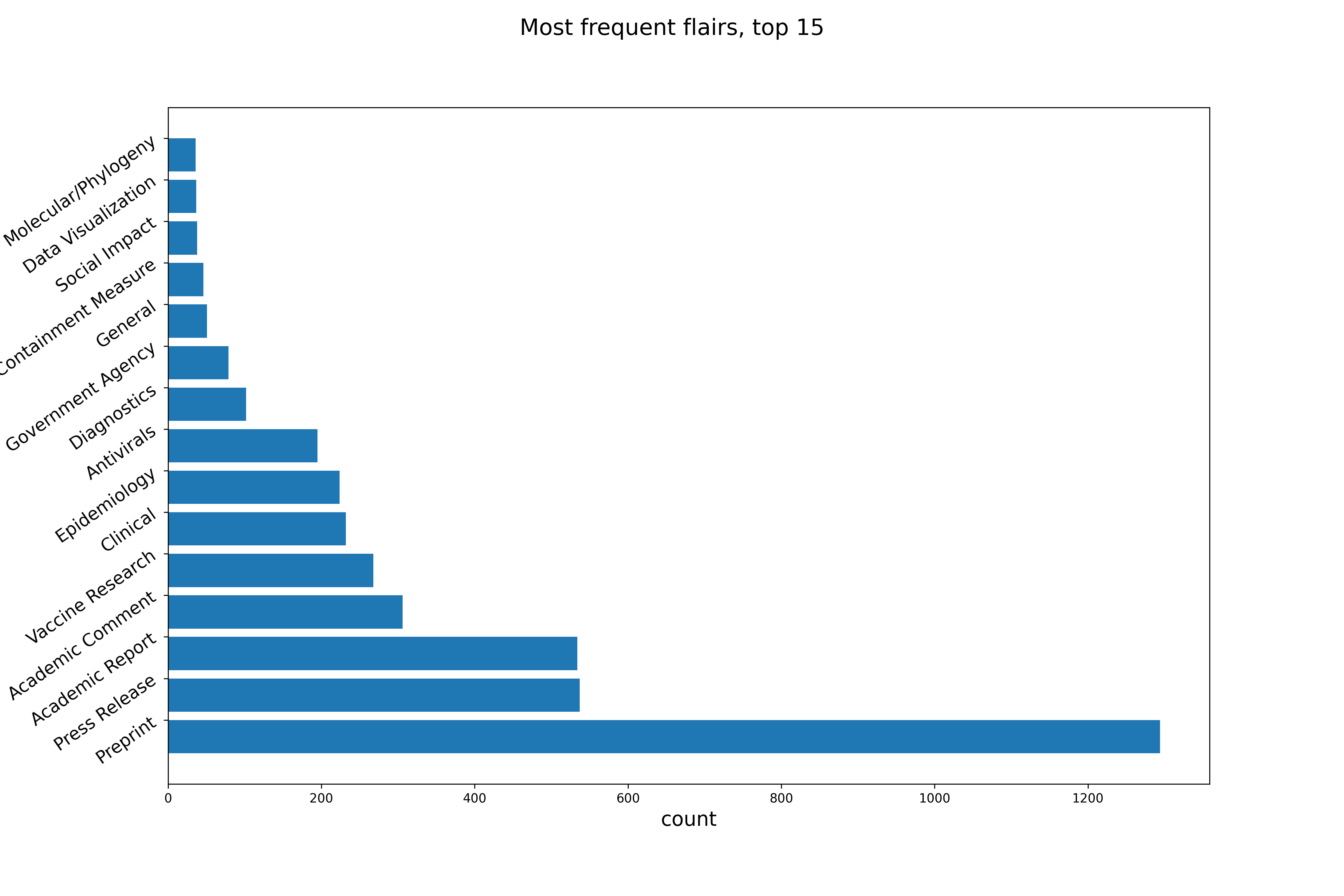}
\caption{\label{figure:image4} Most frequent flairs in the dataset.}
\end{figure}

\begin{figure}[htbp]
\centering
\includegraphics[scale=0.5]{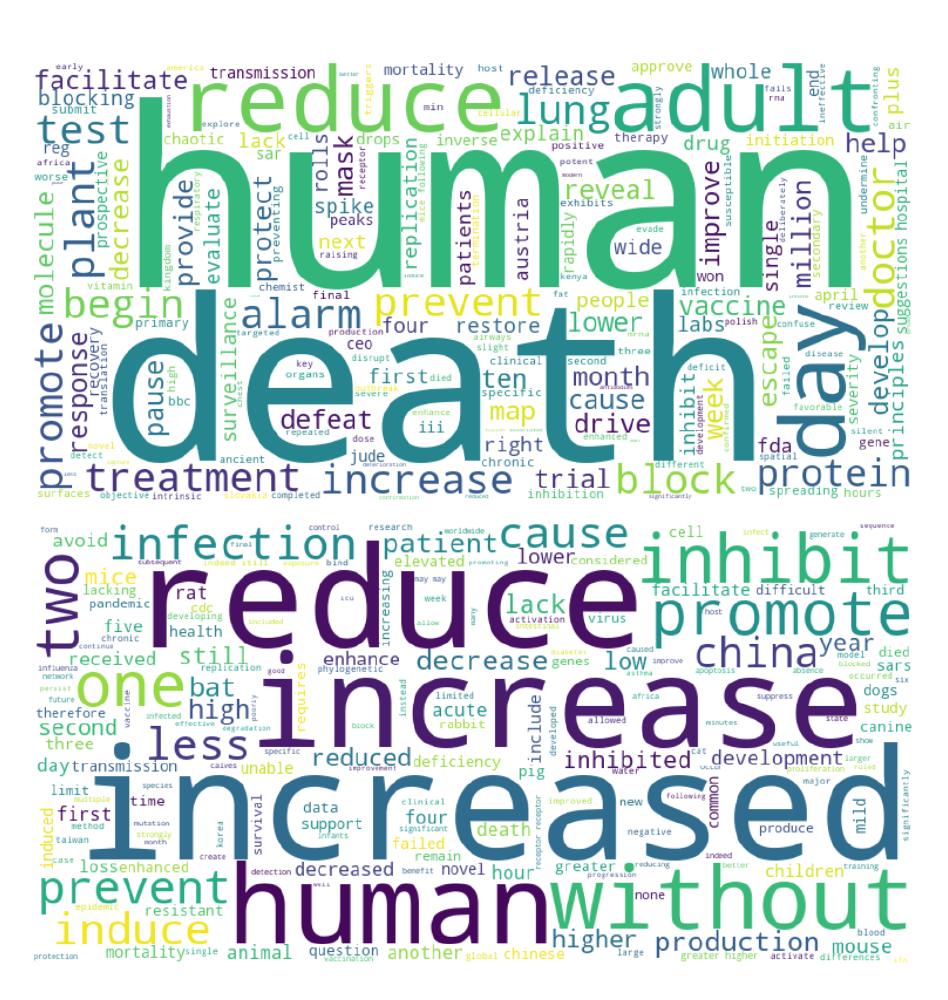}
\caption{\label{figure:cloud} Top image: word cloud of salient words. Bottom image: word cloud of replaced words.}
\end{figure}

\clearpage

\subsection{Word replacement statistics}
\label{appendix:replacement}
Figures below show most frequent salient words as well as most frequent words that were used to replace the salient words (replacement words). POS tags obtained using the flair python library tagger.

\begin{figure}[htbp]
\centering
\includegraphics[scale=0.2]{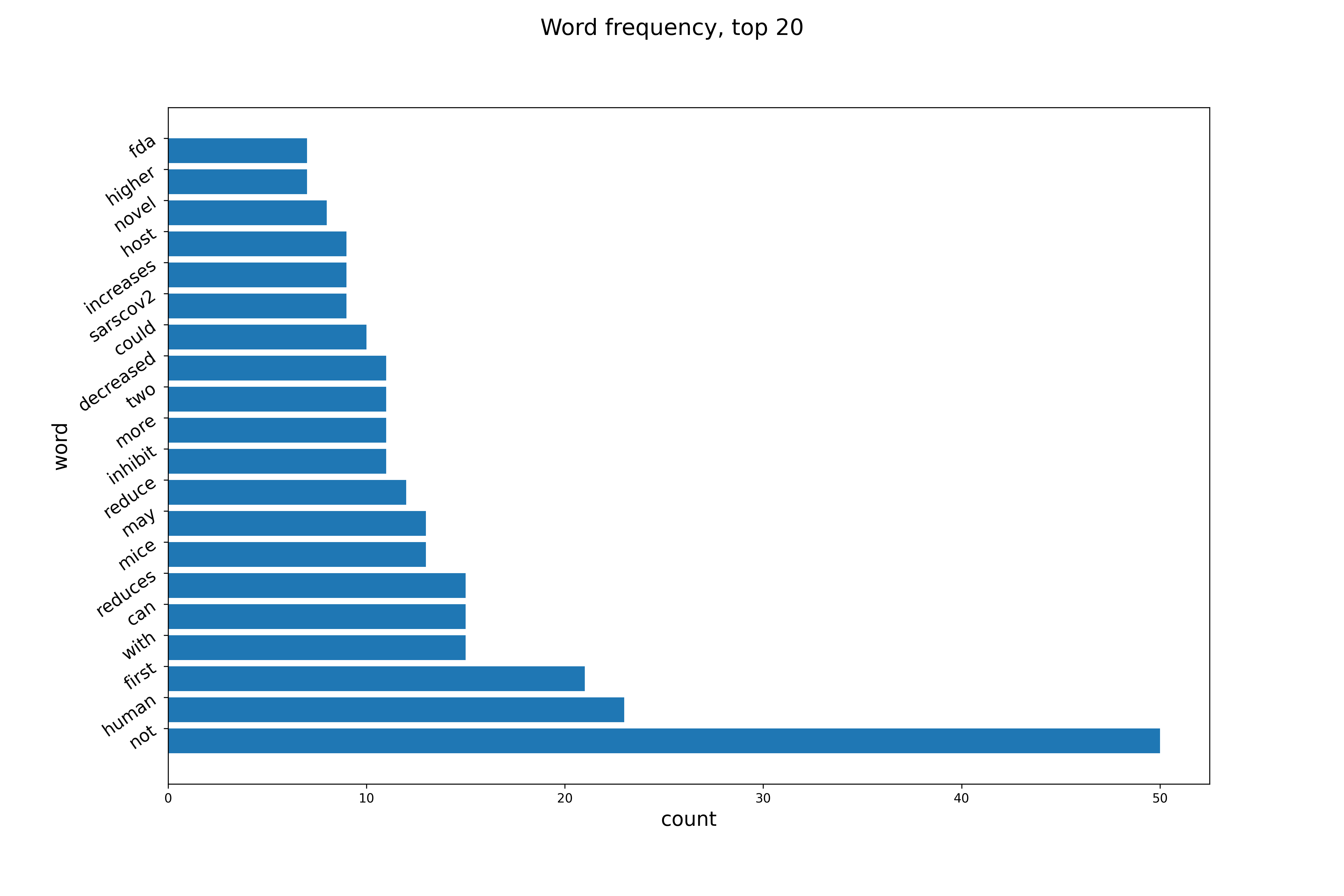}
\caption{Most frequent salient words.}
\label{figure:freqOG}
\end{figure}

\begin{figure}[htbp]
\centering
\includegraphics[scale=0.2]{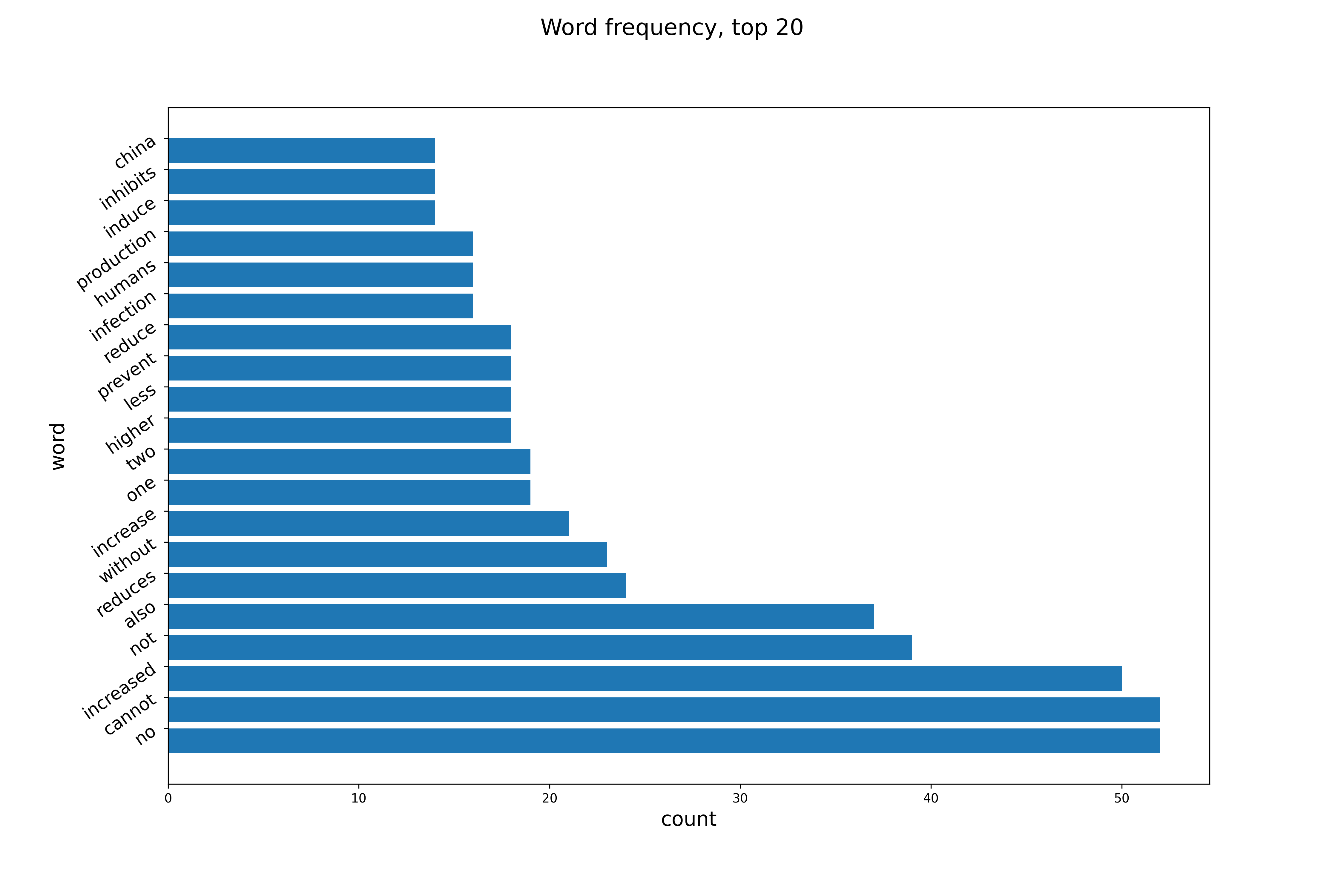}
\caption{Most frequent replacement words.}
\label{figure:freqREPL} 
\end{figure}


\begin{figure}[htbp]
\centering
\includegraphics[scale=0.2]{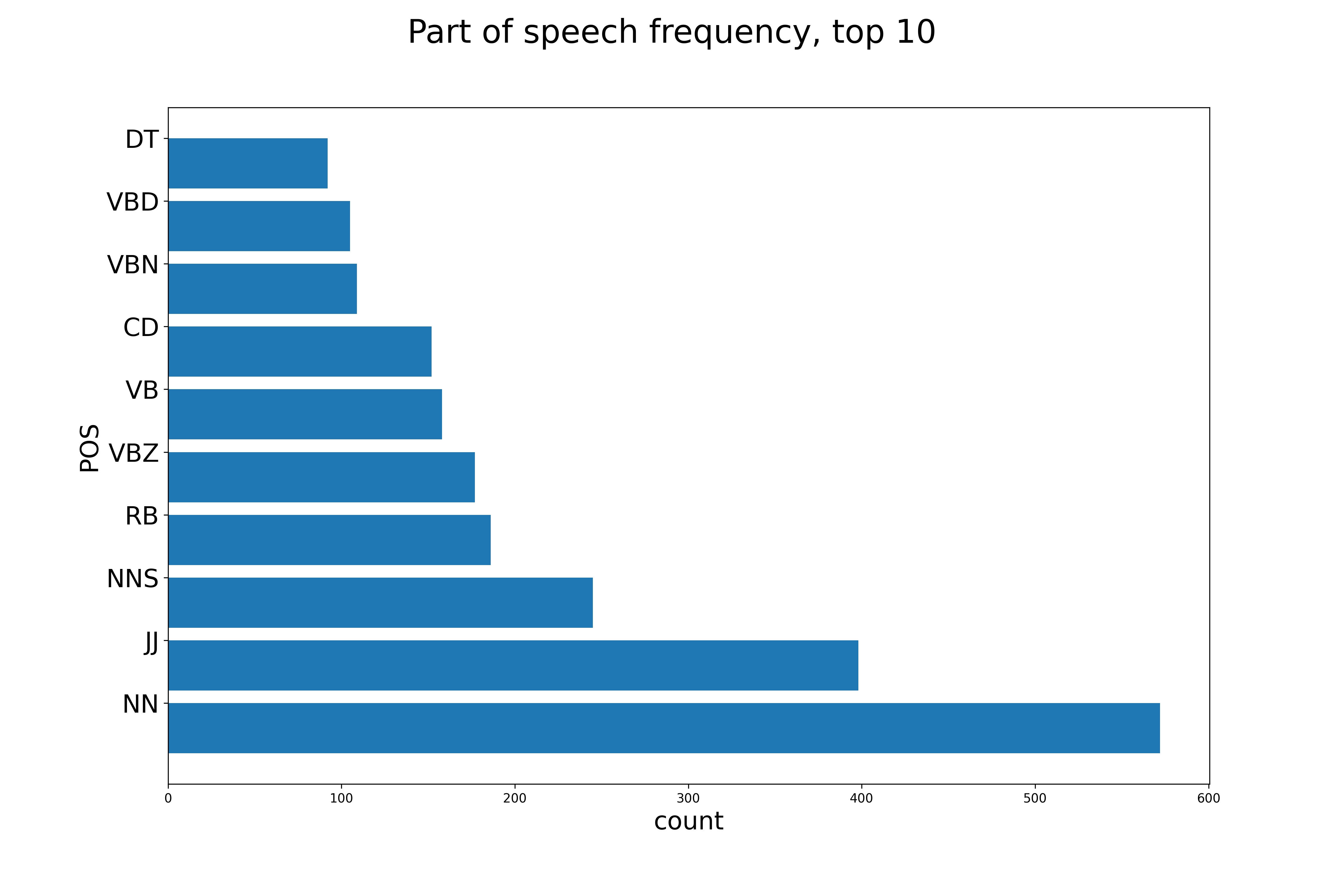}
\caption{Most frequent POS tags of replacement words.}
\label{figure:POSrepl}
\end{figure}

\newpage

\subsection{Alexa threshold}
\label{appendix:alexa}
A boxplot that helped us select the 50,000 Alexa siterank threshold. The plot shows site ranks for 2K initially scraped claims. Outliers (points outside of the whiskers of the plot) are all above the 50,000 threshold.

\begin{figure}[htbp]
\centering
\includegraphics[scale=0.5]{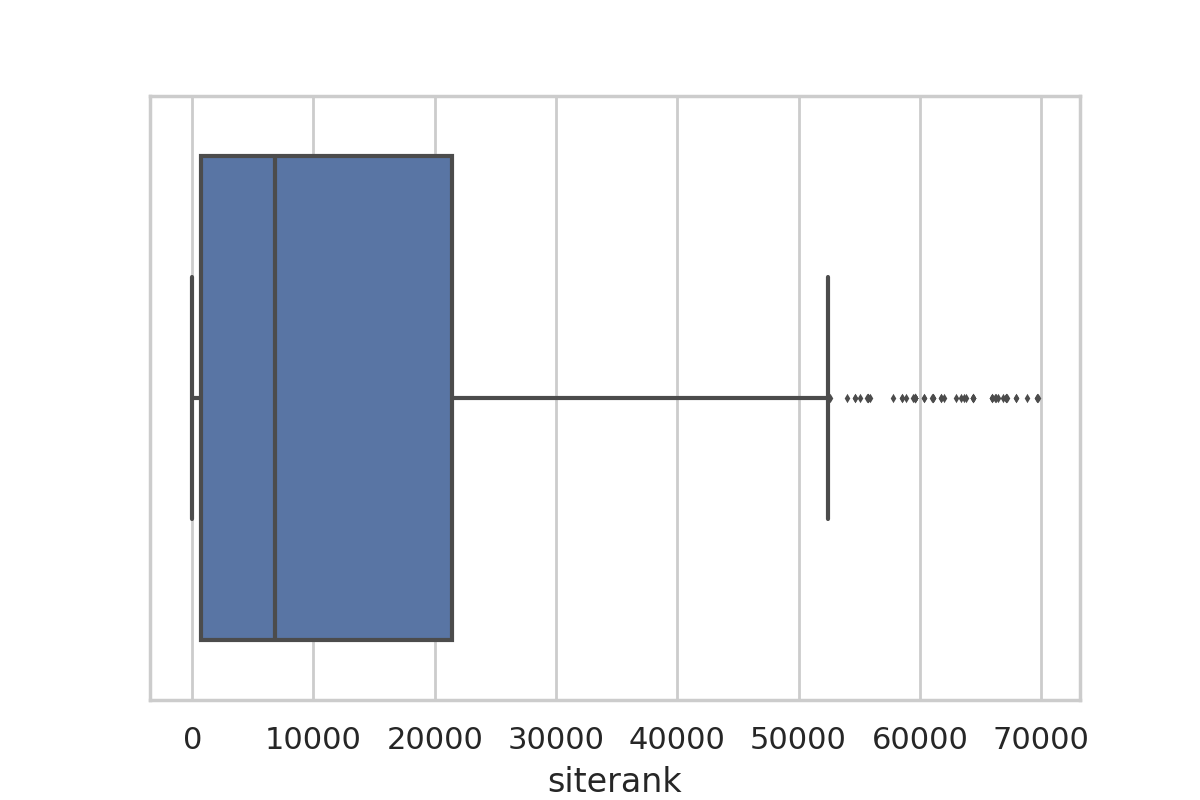}
\caption{\label{figure:siterank} Alexa Site Rank boxplot.}
\end{figure}

\end{document}